\documentclass[a4,twocolumn]{article}

\usepackage{authblk}
\usepackage{moreverb,url}
\usepackage[pdftex]{graphicx}  
\usepackage[numbers]{natbib}
\usepackage{breqn}
\usepackage[usenames,dvipsnames]{xcolor}
\usepackage{caption}
\usepackage{adjustbox}
\usepackage{rotating} 
\usepackage{tabularx}
\usepackage{tabu}
\usepackage{multirow}
\usepackage{subfig}
\usepackage{pgfplots}
\usepackage{verbatim}
\usepackage{tikz}
\usepackage{tikz-qtree}
\usepackage{tikz-dependency}
\usepgfplotslibrary{groupplots}
\usepackage{makecell}
\usepackage[LAE,T1]{fontenc}
\usepackage[utf8]{inputenc}
\usepackage[main=english,farsi,french]{babel}

\makeatletter
\newcommand\footnoteref[1]{\protected@xdef\@thefnmark{\ref{#1}}\@footnotemark}
\makeatother

\newcommand{\ie}{i.e.,~}
\newcommand{\virgolette}[1]{``#1''}

\begin{document}

\title{\textbf{Cross-Lingual Adaptation Using Universal Dependencies}}

\author{\textbf{Nasrin Taghizadeh and Heshaam Faili}}

\affil{School of Electrical and Computer Engineering, \\College of Engineering, University of Tehran, Tehran, Iran\\ \{nsr.taghizadeh, hfaili\}@ut.ac.ir}

\date{}
\maketitle

\begin{abstract}
We describe a cross-lingual adaptation method based on syntactic parse trees obtained from the Universal Dependencies (UD), which are consistent across languages, to develop classifiers in low-resource languages. The idea of UD parsing is to capture similarities as well as idiosyncrasies among typologically different languages. In this paper, we show that models trained using UD parse trees for complex NLP tasks can characterize very different languages. We study two tasks of paraphrase identification and semantic relation extraction as case studies. Based on UD parse trees, we develop several models using tree kernels and show that these models trained on the English dataset can correctly classify data of other languages e.g. French, Farsi, and Arabic. The proposed approach opens up avenues for exploiting UD parsing in solving similar cross-lingual tasks, which is very useful for languages that no labeled data is available for them.

\end{abstract}

\section{Introduction}

% what is UD?

Universal Dependencies (UD) \cite{McDonaldNivreQuirmbach-BrundageEtAl2013,straka2015parsing,nivre2017universal} is an ongoing project aiming to develop cross-lingually consistent treebanks for different languages. UD provided a framework for consistent annotation of grammar (parts of speech, morphological features, and syntactic dependencies) across different human languages\footnote{\label{UD-site}{\url{https://universaldependencies.org/introduction.html}}}. The annotation schema relies on Universal Stanford Dependencies \cite{de2014universal} and Google Universal POS tags \cite{petrov2011universal}. The general principle is to provide universal annotation; meanwhile, each language can add language-specific relations to the universal pool when necessary. 

% importance of UD, % importance of cross-language learning 

The main goal of UD project is to facilitate multi-lingual parser production and cross-lingual learning\footnoteref{UD-site}. Cross-lingual learning is the task of gaining advantages from high-resource languages in terms of annotated data to build a model for low-resource languages. This paradigm of learning is now an invaluable tool for improving the performance of natural language processing in low-resource languages.

% using UD in cross-language learning

Based on the universal annotations of the UD project, there are several works on cross-lingual tasks. Most of them focus on grammar-related tasks such as POS tagging \cite{kim-etal-2017-cross} and dependency parsing \cite{guo-etal-2015-cross,tiedemann2016synthetic,zeman2018conll}. In this paper, we are going to study the effectiveness of UD in making cross-lingual models for more complex tasks such as semantic relation extraction and paraphrase identification. To the best of our knowledge, no work was done on the application of UD annotations in the mentioned tasks.

% Foundomental assuption of UD

Universal dependencies approach for cross-lingual learning is based on the fact that UD captures similarities as well as idiosyncrasies among typologically different languages. The important characteristic of UD annotations is that although the  UD parse trees of parallel sentences in different languages may not be completely equivalent, they have many similar sub-trees, in the sense that at least core parts of trees are equal \cite{nivre-2016-universal}. 

%So far, different approaches for cross-language learning have been proposed.
% Adversarial learning \cite{joty-etal-2017-cross}

In this paper, we study two cross-lingual tasks: semantic relation extraction and paraphrase identification. The former is the task of identifying semantic connections between entities in a sentence; while the training and test data are in different languages. The latter is defined to determine whether two sentences are paraphrase or not; while the training' pairs of sentences are in a different language from the test data.

% motivate why tree-based models, specially tree kernels?

To employ similarities of UD trees of different languages to train cross-lingual models, we propose to use syntactic based methods which ideally can deal with parsing information of data. We found that tree kernels allow to estimate the similarities among texts directly from their parse trees. They are known to operate on dependency parse trees and automatically generate robust prediction models based on the similarities of them. We have made parallel dataset for each task and presented the cross-lingual variant of kernel functions for them. Evaluation by the parallel test data reveals that the accuracy of models trained by a language and tested on the other languages get close to mono-lingual when the syntactic parsers are trained with UD corpora. This suggests that syntactic patterns trained on the UD trees can be invariant with respect to very different languages. 

To compare the proposed approach with the cross-lingual variant of neural models, we employed several state-of-the-art deep networks and equipped them with pre-trained bi-lingual word embeddings. English training data are fed into the networks, which create a mapping between the input and output values. Then test set is given to the trained network. Results show that the tree-based models outperform end-to-end neural models in cross-lingual experiments.

Moreover, we employed Tree-LSTM network \cite{tai-etal-2015-improved} with UD parse trees, which is capable to produce semantic representation from tree-ordered input data. Tree-LSTM doesn't directly deal with syntactic features of the input sentence, rather it processes the input tokens in order of placing in a tree, e.g. from bottom to up or vice versa. Experiments show superiority of Tree-LSTM trained by UD trees over sequential models like LSTM in cross-lingual evaluations.

This paper is organized as follows: Section \ref{sec:ud-high-level-desc} describes how UD approach allows to capture similarities and differences across diverse languages. Section \ref{sec:kernel} presents tree-based models for cross-lingual learning of PI and RE tasks. Section \ref{sec:experiment} presents an empirical study on cross-lingual learning using UD. Finally Section \ref{sec:analysis} gives the analysis and conclusion remarks.

\section{Transfer Learning via Universal Dependencies}
\label{sec:ud-high-level-desc}

The Universal Dependencies project aims to produce consistent dependency treebanks and parsers for many languages \cite{McDonaldNivreQuirmbach-BrundageEtAl2013,straka2015parsing,nivre2017universal}. The most important achievements of the project are the cross-lingual annotation guidelines and sets of universal POS and the grammatical relation tags. Consequentially many treebanks have been developed for different languages. The general rule of UD project is to provide a universal tag set; however each language can add language-specific relations to the universal pool or omit some tags. 

To capture similarities and differences across languages, UD uses a representation consisting of three components: (i) dependency relations between lexical words; (ii) function words modifying lexical words; and (iii) morphological features associated with words \cite{nivre-2016-universal}. 

The underlying principle of the syntactic annotation schema of the UD project is that dependencies hold between content words, while function words attach to the content word that they further specify \cite{de2014universal}. There is an important difference between UD schema and Stanford Typed Dependencies (STD) \cite{de2008stanford} as the STD schema chooses function words as heads: prepositions in prepositional phrases, and copula verbs that have a prepositional phrase as their complement.

\begin{figure*}
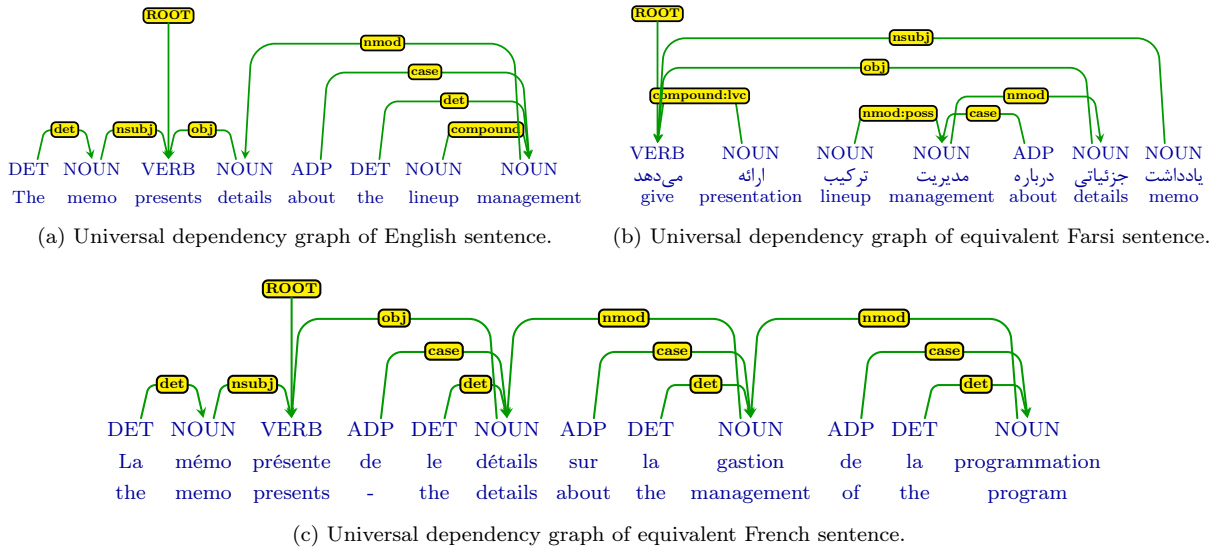

\centering{
\footnotesize
\subfloat[Universal dependency graph of English sentence.\label{fig:ud-en}]
{
\resizebox{\columnwidth}{!}{
\begin{dependency}[theme=brazil]%[text only label,label style={above}]
\begin{deptext}[column sep=.1cm, row sep = 0.1cm]
DET \& NOUN \& VERB \& NOUN \& ADP \& DET \& NOUN \& NOUN \\
The \& memo  \& presents \& details \& about \& the  \& lineup \& management \\
\end{deptext} 
\depedge{1}{2}{det}
\depedge{2}{3}{nsubj}
\depedge{4}{3}{obj} 
\deproot[edge unit distance=5ex]{3}{ROOT} 
\depedge{5}{8}{case}
\depedge{6}{8}{det}
\depedge{7}{8}{compound}
\depedge{8}{4}{nmod}
%\depedge[edge unit distance=3ex]{9}{3}{punct}
\end{dependency}
}
}
\subfloat[Universal dependency graph of equivalent Farsi sentence. \label{fig:ud-fa}]
{
\resizebox{\columnwidth}{!}{
\begin{dependency}[theme=brazil]%[text only label,label style={above}]
\begin{deptext}[column sep=.1cm, row sep =0.1]
VERB \& NOUN \& NOUN \& NOUN \& ADP  \& NOUN \& NOUN \\
\FR{می‌دهد} \& \FR{ارائه} \&  \FR{ترکیب} \& \FR{مدیریت} \& \FR{درباره} \& \FR{جزئیاتی} \& \FR{یادداشت} \\
give \& presentation \& lineup \& management \& about \& details \& memo \\
\end{deptext}
%\depedge{1}{2}{PUNCT} 
\deproot[edge unit distance=4.5ex]{1}{ROOT}
\depedge[edge unit distance=6ex]{2}{1}{compound:lvc}
\depedge{3}{4}{nmod:poss}
\depedge[edge unit distance=3ex]{4}{6}{nmod}
\depedge{5}{4}{case}
\depedge[edge unit distance=1.94ex]{6}{1}{obj}
\depedge[edge unit distance=2.25ex]{7}{1}{nsubj} 
\end{dependency}
} % end of resizebox
} % end of subfloat

\subfloat[Universal dependency graph of equivalent French sentence. \label{fig:ud-fr}]
{
\resizebox{1.7\columnwidth}{!}{
\begin{dependency}[theme=brazil]%[text only label,label style={above}]
\begin{deptext}[column sep=.1cm, row sep = 0.1cm]
DET \& NOUN \& VERB \& ADP \& DET \& NOUN  \& ADP \& DET \& NOUN \& ADP \& DET \& NOUN\\
{La} \& {m\'emo} \& {pr\'esente} \& {de} \& {le} \& {d\'etails} \& {sur} \& {la} \& {gastion} \& {de} \& {la} \& {programmation} \\
the \& memo \& presents \& - \& the \& details \& about \& the \& management \& of \& the \& program\\
\end{deptext} 
\depedge{1}{2}{det}
\depedge{2}{3}{nsubj}
\deproot[edge unit distance=4ex]{3}{ROOT} 
\depedge{4}{6}{case}
\depedge{5}{6}{det}
\depedge{6}{3}{obj}
\depedge{7}{9}{case}
\depedge{8}{9}{det}
\depedge{9}{6}{nmod} 
\depedge{10}{12}{case}
\depedge{11}{12}{det}
\depedge{12}{9}{nmod}
%\depedge[edge unit distance=1.5ex]{13}{3}{punct}
\end{dependency}
} % end of resizebox
} % end of subfloat

} %end of centering

\caption{UD annotations for equivalent sentences in English, Farsi, and French. English translation of each word is written below it.}
\label{fig:ud-graph-compare}
\end{figure*}

Although the UD parse graphs of a sentence in different languages may not be completely equal, they have similar core parts. Figure \ref{fig:ud-graph-compare} shows the UD graph of English sentence ``\textit{The memo presents details about the lineup management}" and its translation into French and Farsi. Both the similarities and differences of UD graphs are demonstrated in that figure\footnote{In Farsi, sentences are written from right to left.}. Most of the nodes and edges are similar. Farsi has the language-specific relation ``\texttt{compound:lvc}", which relates the noun part of the compound verb to the verbal part as depicted in Figure \ref{fig:ud-fa}. So far, UD treebanks have been developed for over 70 languages and all of them are freely available for download\footnote{\url{https://github.com/ufal/udpipe}}. UD project released a pipeline, called UDPipe\footnote{\url{http://lindat.mff.cuni.cz/services/udpipe/run.php}}, which is used to train models for UD parsing using the UD treebanks \cite{straka2016udpipe}.

UD parsing and similarity of UD structures in different languages provide facilities to train multi-lingual models. In what follows, we focus on two tasks, paraphrase identification and semantic relation extraction, and present cross-learning models for them.

\section{Cross-Lingual Tree-based Models}
\label{sec:kernel}

To employ UD parsing in cross-lingual learning, there should be a training algorithm that is capable of utilizing similarities of UD parse trees in different languages. Kernel methods such as SVM use a similarity function, which is called kernel function, to assign a similarity score to pairs of data samples. A kernel function $K$ over an object space $X$ is symmetric, positive semi-definite function $K: X \times X \rightarrow [0,\infty)$ that assigns a similarity score to two instances of $X$, where $K(x,y)=\phi(x)\cdot \phi (y)=\sum{\phi_{i}(x)\phi_{i}(y)}$. Here, $\phi(x)$ is a mapping function from the data object in $X$ to the high-dimensional feature space. Using the kernel function, it is not necessary to extract all features one by one and then multiply the feature vectors. Instead, kernel functions compute the final value directly based on the similarity of data examples.

Tree kernels are the most popular kernels for many natural language processing tasks \cite{panyam2018exploiting,filice2015structural}. Tree Kernels compute the number of common substructures between two trees $T_1$ and $T_2$ without explicitly considering the whole fragment space \cite{croce2011structured}. Suppose the set $\mathcal{F}=\{f_1,f_2, \dots, f_{|\mathcal{F}|} \}$ be the tree fragment space and $\mathcal{X}_i(n)$ be an indicator function that is 1 if the $f_i$ rooted at node $n$ and equals to 0, otherwise. Now, tree kernel over $T_1$ and $T_2$ is defined as below \cite{croce2011structured}:
\begin{equation}
K(T_1, T_2) = \sum_{n_1 \in N_{T_1}} \sum_{n_2 \in N_{T_2}} \Delta (n_1,n_2),
\end{equation}
where $N_{T_1}$ and $N_{T_2}$ are the set of nodes of $T_1$ and $T_2$, respectively and 
\begin{equation}
\Delta (n_1,n_2) = \sum_{i=1}^{|\mathcal{F}|}{\mathcal{X}_i(n_1) \mathcal{X}_i(n_2)},
\end{equation}
which shows the number of common fragments rooted in $n_1$ and $n_2$ nodes. Different tree kernels vary in their definition of $\Delta$ function and fragment type. 

There are three important characterizations of fragment type \cite{nguyen2009convolution}: SubTree, SubSet Tree and Partial Tree. A SubTree is defined by taking a node of a tree along with all its descendants. SubSet Tree is more general and does not necessarily contain all of the descendants. Instead, it must be generated by utilizing the same grammatical rule set of the original trees. A Partial Tree is more general and relaxes SubSet Tree's constraints. Some popular tree kernels are SubSet Tree Kernel (SST), Partial Tree Kernel (PTK) \cite{moschitti2006efficient} and Smoothing Partial Tree Kernel (SPTK) \cite{croce2011structured}. In the next section, we employ the tree kernels along with UD parse trees for solving cross-lingual tasks.

\subsection{Cross-Lingual Paraphrase Identification}
\label{sec:cross-pi}
Paraphrase Identification (PI) is the task of determining whether two sentences are paraphrase or not. It is considered a binary classification task. The best mono-lingual methods often achieve about 85\% accuracy over this corpus \cite{filice2015structural,wang-etal-2016-sentence}. Filice et al. \cite{filice2015structural} extended the tree kernels described in the previous section to operate on text pairs. The underlying idea is that this task is characterized by several syntactic/semantic patterns that a kernel machine can automatically capture from the training material. We can assess a text pair as a paraphrase if it shows a valid transformation rule that we observed in the training data. The following example can clarify this concept. A simple paraphrase rewriting rule is the active-passive transformation, such as in \virgolette{\textit{Federer beat Nadal}} and \virgolette{\textit{Nadal was defeated by Federer}}. The same transformation can be observed in other paraphrases, such as in \virgolette{\textit{Mark studied biology}} and \virgolette{\textit{Biology was learned by Mark}}. Although these two pairs of paraphrases have completely different topics, they have a very similar syntactic structure. %, as Figure \ref{fig:inter-pair} illustrates. 

%\begin{figure}[!ht]
%\centering
%
%\caption{UD trees of two pairs of paraphrases applying the same active-passive transformation rule.}
%\label{fig:inter-pair}
%\end{figure}

Tree kernel combinations can capture this inter-pair similarity and allow a learning algorithm such as SVM to learn the syntactic-semantic patterns characterizing valid paraphrases. Given a tree kernel $TK$ and text pairs $p_i = (i_1, i_2)$, the best tree kernel combination for the paraphrase identification task described in \cite{filice2015structural} is the following:

\begin{small}
\begin{dmath}
SM_{TK} ( p_a, p_b ) = softmax \Big ( TK(a_1,b_1)TK(a_2, b_2), TK(a_1,b_2)TK(a_2,b_1) \Big ) \end{dmath}
\end{small}

\noindent where \emph{softmax}$(x_1,x_2)= \frac{1}{m}  \log \left(e^{m x_1} + e^{m x_2}\right)$ is a simple function approximating the max operator\footnote{m=100 is accurate enough.}, which cannot be directly used in kernel formulations, as it can create non valid kernel functions. In this kernel combination the two different alignments between the trees of the two pairs are tried and the best alignment is chosen. This allows to exploit the inherent symmetry of the Paraphrase Identification task (\ie if $a$ is a paraphrase of $b$, it also implies that $b$ is a paraphrase of $a$).

When we adopt the universal dependencies, different languages have a common formalism to represent text syntax, and tree kernels, that mostly operate at a syntactical level, can still provide reliable similarity estimations, \ie $SM_{TK}(p_a, p_b)$ can work even if $p_a$ and $p_b$ have different languages. This allows operating in a cross-lingual setting. For instance, we can use a model trained on a high-resource language for classifying textual data of a poor-resource language. In addition to the syntactic similarity evaluation, the PTK and SPTK which are used in the $SM_{TK}$ formulation also perform a lexical matching among the words of the trees to be compared.

\subsection{Cross-Lingual Semantic Relation Extraction}
\label{sec:cross-re}
Relation Extraction (RE) is defined as the task of identifying semantic relations between entities in a text. The goal is to determine whether there is a semantic relation between two given entities in a text, and also to specify the type of relationship if present. RE is an important part of Information Extraction \cite{taghizadeh2020nsurl2019}. Relation extraction methods often focus on the Shortest Dependency Path (SDP) between entities \cite{le2019richer}. However, there are some crucial differences between UD annotation principles and others parse formalisms that causes us to reconsider SDP of UD trees. 

Considering the sentence: ``\textit{The most common $[$audits$]_{e1}$ were about $[$waste$]_{e2}$ and recycling}", there is a Message-Topic relation between $e1$ and $e2$. The most informative words of the sentence for the relation are ``\textit{were}" and ``\textit{about}"; while the other words of the sentence can be ignored and the same relation is still realized. It is a crucial challenge of relation extraction methods that important information may appear at any part of the sentence. Most previous works assume that the words lying in the window surrounding entities are enough to extract the relation governing entities \cite{hashimoto-EtAl:2015:CoNLL,bokharaeian2017snpphena}. However, words of a sentence are often reordered when the sentence is translated into other languages. 
%For example words around two entities in English sentence may go far from entities in equivalent Farsi sentence. Especially when the verb of the sentence decides the type of semantic relation, reordering of words causes that verb go to the end of the sentence in SOV languages; while it was near to entities in SVO languages. 
Therefore, using words in the window surrounding entities may result in an accurate model for mono-lingual experiments, but not necessarily for cross-lingual ones. 

%It is shown that the words lying in the shortest dependency path between two entities have decisive effect on the semantic relation of entities \cite{moschitti2006efficient,croce2011structured,gamallo2012dependency}. 
Regarding UD parsing, there are several significant differences between universal annotation schema and other schemas for dependency parsing. Two main differences are related to prepositions and copula verbs. According to the UD annotation guidelines, prepositions are attached to the head of a nominal, and copula verbs are attached to the head of a clause. However in other schemas, prepositions are often the root of the nominal, and the clause is attached to the copula. 

\begin{figure}
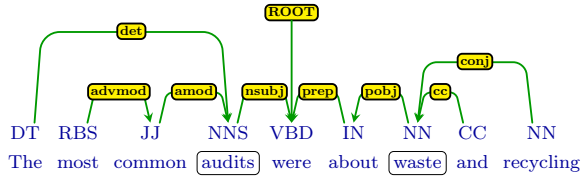
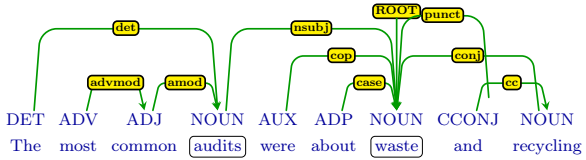

\centering{
\footnotesize
\subfloat[Non-universal dependency parse tree. \label{fig:non-ud-rel}]
{
\resizebox{\columnwidth}{!}{
\begin{dependency}[theme=brazil]%[text only label,label style={above}]
\begin{deptext}[column sep=.1cm, row sep = 0.1cm]
DT \& RBS \& JJ \& NNS \& VBD \& IN \& NN \& CC \& NN \\
The \& most  \& common \& audits \& were \& about  \& waste \& and \& recycling \\
\end{deptext}
\depedge{1}{4}{det} 
\depedge{2}{3}{advmod}
\depedge{3}{4}{amod}
\depedge{4}{5}{nsubj}
\deproot{5}{ROOT}
\depedge{6}{5}{prep}
\depedge{7}{6}{pobj}
\depedge{8}{7}{cc} 
\depedge{9}{7}{conj} 

% group on 2nd row, from word 1 to word 2, labeled "a0"
\wordgroup{2}{4}{4}{a0}
\wordgroup{2}{7}{7}{a1}

\end{dependency}
} % end of resizebox
} % end of subfloat

\subfloat[Universal dependency parse tree.\label{fig:ud-rel}]
{
\resizebox{\columnwidth}{!}{
\begin{dependency}[theme=brazil]%[text only label,label style={above}]
\begin{deptext}[column sep=.1cm, row sep = 0.1cm]
DET \& ADV \& ADJ \& NOUN \& AUX \& ADP \& NOUN \& CCONJ \& NOUN \\
The \& most  \& common \& audits \& were \& about  \& waste \& and \& recycling \\
\end{deptext} 
\depedge{1}{4}{det}
\depedge{2}{3}{advmod}
\depedge{3}{4}{amod}
\depedge{4}{7}{nsubj} 
\depedge{5}{7}{cop}
\depedge{6}{7}{case}
\deproot{7}{ROOT}
\depedge{8}{9}{cc}
\depedge{9}{7}{conj}
\depedge{10}{7}{punct}

% group on 2nd row, from word 1 to word 2, labeled "a0"
\wordgroup{2}{4}{4}{a0}
\wordgroup{2}{7}{7}{a1}

\end{dependency}
}
}
} %end of centering
\caption{Two different parse trees for the sentence ``The most common $[$audits$]_{e1}$ were about $[$waste$]_{e2}$ and recycling".}
\label{fig:dep-graph-compare}
\end{figure}

Figure~\ref{fig:non-ud-rel} shows the parse tree of the example: ``\textit{The most common $[$audits$]_{e1}$ were about $[$waste$]_{e2}$ and recycling}". The tree is produced by the ARK\footnote{\url{http://demo.ark.cs.cmu.edu/parse}} parser, which does not follow universal schema. As mentioned before, ``\textit{were}" and ``\textit{about}" lie on the SDP between $e1$ and $e2$. However, considering the UD parse tree depicted in Figure~\ref{fig:ud-rel}, there is no word in the SDP; while both ``\textit{were}" and ``\textit{about}" are attached to $e2$. As a result, we propose that the words which are dependent on the entities be considered to be the informative words in addition to the SDP's words. We use these words for making a cross-lingual model.

%In what follows, we explain how lexical features including informative words of a sentence, and structural features, which are obtained from parse trees, are incorporated in the cross-lingual model. We propose a cross-lingual variant of SVM and neural networks algorithms. 

%------------------------ 
Kernel functions have several interesting characteristics. The combination of kernel functions in a linear or polynomial way results in a valid kernel function \cite{moschitti2012state}. Composite kernel functions are built on individual kernels; each of them captures part of the features of a data object. Tree kernels capture the data's syntactic structure, while a word sequence kernel considers the words of a sequence in a particular order. To define a cross-lingual kernel, we have adopted the composite kernel used by the Nguyen et al. \cite{nguyen2009convolution}:
\begin{equation}
\label{eq:ck}
CK = \alpha . K_{SST} + (1-\alpha).(K_{P-e}+K_{PT})^2,
\end{equation}
where $K_{P-e}$ is a polynomial kernel. Its base kernel is an entity kernel ($K_E$), which is applied to an entity-related feature vector consisting of (named) entity type, mention type, headword, and POS tag.
%\begin{equation}
%K_{P-e} = (1+K_E)^2,
%\end{equation}
%\begin{equation}
%K_E = \sum_{i=1,2}{K_L(R_1.E_i, R_2.E_i)},
%\end{equation}
%wherer $K_L$ is linear kernel. 
$K_{SST}$ is the Sub-Set Tree (SST) kernel, which is applied to the Path-Enclosed Tree (PET) of the constituency tree structure. PET is the smallest common subtree including the two entities \cite{moschitti2004study,taghizadeh2018cross}. $K_{PT}$ is the Partial Tree kernel \cite{moschitti2006efficient}, which is applied to the dependency-based tree structures. Parameter $\alpha$ weighs the kernels.

To incorporate the most informative words of the sentence into the model, the feature vector $V_o$ is defined similarly to the work of Hashimoto et al. \cite{hashimoto-EtAl:2015:CoNLL}. They proposed concatenating these vectors to make the $V_o$: the vector representing $e1$, the vector representing $e2$, the average of vectors representing words between two entities, the average of vectors representing words in a window before $e1$, and the average of vectors representing words in a window after $e2$. 

Since $V_o$ is defined based on the position of words in the sentence and thus is not necessary a cross-lingual consistent feature vector, we propose to define feature vector $V_{ud}$ by concatenating these vectors: the vector representing $e1$, the vector representing $e2$, the average of vectors representing words in the shortest path between two entities (instead of words between $e1$ and $e2$), the average of vectors representing words dependent to $e1$ (instead of words before $e1$), and the average of vectors representing words dependent to $e2$ (instead of words after $e2$). 
$V_{ud}$ is cross-lingually consistent provided that the words are picked up from UD parse trees and represented by multi-lingual embeddings.

Based on the $CK$ defined in formula \ref{eq:ck} and the feature vectors $V_o$ and $V_{ud}$, the following composite kernels are proposed:
%\begin{equation}
%\label{eq-ck0}
%CK_0 = (K_{P-e}+K_{PT})^2,
%\end{equation}

\begin{equation}
\label{eq-ck1}
CK_1 = \alpha . K_{SST} + (1-\alpha).(K_{P-o}+K_{PT})^2,
\end{equation}
where $K_{P-o}$ is polynomial kernel applied on a feature vector $V_o$.

\begin{equation}
\label{eq-ck2}
CK_2 = (K_{P-ud}+K_{PT})^2,
\end{equation}
where $K_{P-ud}$ is polynomial kernel applied on a feature vector $V_{ud}$. 

\begin{equation}
\label{eq-ck3}
CK_3 = \alpha . K_{SST} + (1-\alpha).(K_{P-ud}+K_{PT})^2
\end{equation}

Constituency parsing of a sentence in a language depends on the syntactic rules governing the position of words. In general, constituency parse trees of a sentence in different languages are different. So, the constituency tree should not be involved in the cross-lingual model. Here, $CK_2$ is our proposed kernel, which is used for CL-RE. However, $CK_1$ and $CK_3$ can also be used for cross-lingual experiments subject to the similarity of syntactic parsing of the source and target languages. 

%However, $CK_1$ is defined to compare the accuracy of the $CK_1$ and $CK_3$ in mono-lingual setting. Kernel $CK_3$ is also defined to be used in the competitor method; when the training data is translated by MT and the model is trained directly on the translated data. In this case, training and test data have the same language and so constituency parsing could be involved.

%To detect the relation between entities, corresponding nodes must be specified in the tree somehow. Therefore, two extra nodes, namely $E1$ and $E2$ denoting first and the second entities, are added to the constituency tree, above of the target entity nodes similar to the work of Nguyen et al. \cite{nguyen2009convolution}. 

SST kernel works only on the constituency trees and not on the dependency trees \cite{moschitti2006efficient}. Therefore, for evaluating the similarity of dependency trees, PT kernel is used. The PT kernel cannot process labels on the edges; so dependency trees are converted to the Lexical Centered Tree (LCT) format \cite{croce2011structured} and then PT kernel is applied on the transformed trees. In LCT format, the lexical is kept at the center and the other information related to that lexical, such as POS tag and grammatical relation, is then added as its children. %Figure \ref{fig:lct-example} shows the sentence of Figure \ref{fig:ud-rel} in LCT format. Using the proposed composite kernels, the training and test data are presented in the same space.

%-------------------------

MultiWord Expression (MWE) is a lexeme made up a sequence of two or more lexemes as each lexeme has its own meaning, but the meaning of the whole expression cannot (or at least can only partially) be computed from the meaning of its parts. MWE displays lexical, syntactic, semantic, pragmatic and/or statistical idiosyncrasies \cite{indurkhya2010handbook}. The nature of MWE leads us to deal with the whole lexemes as a word. Fortunately, MWE can be identified from the parse tree. There are three types of dependency relations for MWE in UD parsing: \texttt{flat}, \texttt{fixed}, and \texttt{compound}. According to UD guidelines, the \texttt{flat} relation is used for exocentric (headless) semi-fixed MWEs like names (\textit{Walter Burley Griffin}) and dates (\textit{20 November}). The \texttt{fixed} relation applies to completely fixed grammaticized (function word-like) MWE (like \textit{instead of, such as}), whereas \texttt{compound} applies to endocentric (headed) MWE (like \textit{apple pie}).

\begin{figure}
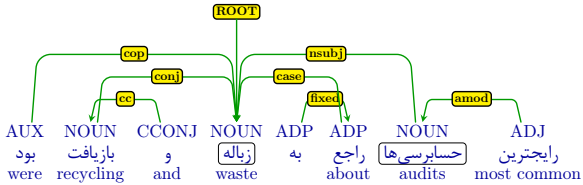

\centering
\resizebox{\columnwidth}{!}{
\begin{dependency}[theme=brazil]%[text only label,label style={above}]
\begin{deptext}[column sep=.1cm, row sep =0.1]
AUX \& NOUN \& CCONJ \& NOUN \& ADP \& ADP  \& NOUN \& ADJ \\
\FR{بود} \& \FR{بازیافت} \& \FR{و} \&  \FR{زباله} \& \FR{به} \& \FR{راجع} \& \FR{حسابرسی‌ها} \& \FR{رایجترین} \\
were \& recycling \& and \& waste \& \& about  \& audits \& most common \\
\end{deptext}
\depedge{1}{4}{cop} 
\depedge{2}{4}{conj}
\depedge{3}{2}{cc}
\deproot[edge unit distance=4ex]{4}{ROOT}
\depedge{5}{6}{fixed}
\depedge{6}{4}{case}
\depedge{7}{4}{nsubj}
\depedge{8}{7}{amod}
\wordgroup{2}{4}{4}{e1}
\wordgroup{2}{7}{7}{e2}
\end{dependency}
}
\caption{Example of MWE in a Farsi sentence. The nodes 5th and 6th from the left make an MWE, while only the 6th node is dependent on the entity.}
\label{fig:mwe}
\end{figure}

To produce feature vector $V_{ud}$, it is better to treat MWE as single words, especially MWE with \texttt{fixed} relations between parts because considering each part of the MWE separately and averaging their embedding vectors may result in a meaningless vector. This point matters when the words of low-resource languages are first translated into other languages and then presented by an embedding of that language. Therefore, the procedure of producing feature vector $V_{ud}$ should be modified with a simple heuristic: every node of the UD tree within the shortest path between two entities or dependent to $e1$ or $e2$ which have a child node with \texttt{fixed} dependency type is considered with its child as one word. If the child has also a child with a \texttt{fixed} dependency, all of them are considered as one word. For example, Figure \ref{fig:mwe} shows the UD tree of a Farsi sentence which is the translation of the English sentence in Figure \ref{fig:dep-graph-compare}. Entities are distinguished from other nodes by putting a circle around them. The 5th and 6th nodes from the left make a multiword expression that means ``about". Applying the above heuristic results in them both being considered as a single word and so the correct translation to another language is found. Some other examples of Farsi MWEs are ``\FR{قبل از آن که}/before", ``\FR{در حالی که}/while", ``\FR{به درون}/into", ``\FR{به جز}/except", and ``\FR{بر روی}/on". In French language there are also MWEs, such as ``bien que/although", ``en tant que/as", ``tant de/so many", ``afin de/in order to", ``pr\'es de/near".

Apart from \texttt{fixed}, \texttt{flat}, and \texttt{compound}, there are grammatical relations which are language-specific and show MWE structures \cite{kahane2017multi}. If the target language has language-specific relations, the above heuristic should be applied to them. For example, \texttt{compound:lvc} relation, which is defined for several languages including Farsi, represents the dependence from the noun part to the light verb part of compound verbs. An example of this relation was shown in Figure \ref{fig:ud-fa}. The words ``\FR{ارائه}/presentation" and ``\FR{می‌دهد}/give" together mean ``present".

\section{Experiments}
\label{sec:experiment}

In this section, the experimental analysis of the proposed models is presented. We have implemented the cross-lingual variant of kernel functions for PI and RE tasks as described in section \ref{sec:kernel} and measured the accuracy of models by testing them on the parallel data set. 

The main advantage of the proposed method is that it needs no data of the test language, in the sense that the model trained using the training data of a language, e.g. English, is directly used in the other languages, e.g. Farsi, Arabic, etc. From this point of view, the proposed method can only be compared with those methods that use no data (neither labeled nor un-labeled) of the test language or parallel corpus or machine translators between the training and test languages.

One solution for cross-lingual tasks is to equip the high accurate neural networks proposed for each task with pre-trained multi-lingual word embeddings, without any change in the architecture of the network. Therefore, we re-implemented some deep methods and compared the proposed approach with them for both PI and RE tasks.
% تلاش کردم به گونه بگویم که داور احساس کند ما رقیب دیگری برای مقایسه نداریم.

\subsection{Paraphrase Identification}
For this task, we made a parallel test dataset and implemented PT and SPT kernels and compared the results with two-channel CNN of Wang et al. \cite{wang-etal-2016-sentence}. 

\subsubsection{Construction of Parallel Dataset}

To prepare a multi-language corpus for PI, we employed an existing English corpus with its Arabic translation and made Farsi correspondence. Microsoft Research Paraphrase Corpus (MSRC) \cite{dolan2004unsupervised} mostly used by the researches for English PI task. It contains 4,076 and 1,725 pairs of sentences for the training and test, respectively. This data has been extracted from news sources on the web, and has been annotated by humans whether each pair captures a paraphrase equivalence relationship.

PI relates to the task of Semantic Textual Similarity (STS), in which the goal is to capture the degree of equivalence of meaning rather than making a binary decision. SemEval-2017 task 1\footnote{\url{http://alt.qcri.org/semeval2017/task1/index.php?id=data-and-tools}} put the emphasis on multi-lingual STS \cite{cer-etal-2017-semeval}. They selected 510 pairs from the test part of the MSRC corpus, and translated them into Arabic by Arabic native speakers. All data have been manually tagged with a number from 0 to 5 to show the degree of similarity. 

The Arabic part of the STS dataset of SemEval-2017 is parallel to some parts of the MSRC test corpus. So there is a parallel English-Arabic dataset. Because of the similarity between PI and STS tasks, the dataset of STS can also be used in the PI task, just by converting the scores to 0 or 1. So, the original binary scores of the STS dataset have been retrieved from the MSRC corpus. As a result, a corpus with 510 pairs of English sentences and Arabic translation for PI task is ready. In addition to Arabic translation, we produced correspondence Farsi data by translation of parallel English-Arabic dataset into Farsi by a Farsi native speaker.

%Then a preprocessing step was conducted so that Farsi data could be used in next experiments. The main activities of preprocessing are 1) unifying different written forms of characters \FR{y}/y, \FR{h}/h and \FR{A}/a; and 2) unifying zero-width non-joiner space named pseudo-space over the corpus.

%Although Farsi has only one written form for its alphabet against Arabic, some typist use Arabic keyboard for typing Farsi text. It is necessary to unify different forms of those letters having multiple forms; thus a word that appearing with different written forms of its alphabets would be considered as unique word. 

%Some Farsi words have two parts that could be written with space or pseudo-space between them. The true form is that one has pseudo-space. Ignoring pseudo-space raises problems in analysis of detached morphemes, specially plural morpheme \FR{hA} , post-determiner \FR{Ay}, present/past continuous morpheme \FR{my}, comparative suffix \FR{tr}, and superlative suffix \FR{tryn} \cite{ghayoomi2010study}. To avoid errors in tokenization and so in grammatical relations of the parse trees, the corpus was reviewed for pseudo-spaces and was corrected manually. 

In the experiments, MSRC corpus was divided as follows: 1) the training part of MSRC corpus for training; 2) those data from test part of MSRC, which we don't have their Arabic or Farsi counterpart, for tuning hyper-parameters as development set; and 3) 510 parallel English-Arabic-Farsi from the test part of MSRC for the test. Therefore, our training and test data have 4076 and 510 samples, respectively. Table \ref{tb:msrc-split} shows the statistics of our data.

\begin{table*}
%\begin{adjustbox}{width=\textwidth,center=\textwidth}
\centering
\small{
\begin{tabular}{|c|c|c|c|}
\hline 
 & Training & Test & Development\\ 
\hline 
original split & 4,076 (En) & 1,725 (En) & 0 \\ 
\hline 
our split & 4,076 (En) & 510 (parallel En-Ar-Fa) & 1215 (En) \\ 
\hline 
\end{tabular} 
}
%\end{adjustbox}
\caption{MSRC corpus split}
\label{tb:msrc-split}
\end{table*}

\subsubsection{Tools and Setup}
The classifiers were trained with the C-SVM learning algorithm within KeLP \cite{filice2015kelp}, which is a kernel-based machine learning framework and implemented tree kernels. We employed PT and SPT kernel functions. For evaluating node similarity in SPTK function, we used the same method described in \cite{filice2015structural}: if $n_1$ and $n_2$ are two identical syntactic nodes, $\sigma(n_1,n_2)$ denoted the similarity of $n_1$ and $n_2$ and is equal to 1. If $n_1$ and $n_2$ are two lexical nodes with the same POS tag, their similarity is computed as the cosine similarity of the corresponding vectors in a wordspace. In all other cases $\sigma = 0$. 

English wordspace was generated by using word2vec tool\footnote{\url{https://code.google.com/p/word2vec/}}. In the cross-lingual setup, we need a vocabulary to find the translation of lexical nodes and then compute their similarity in a wordspace. For English-Arabic experiments, we used Almaany dictionary\footnote{\url{http://www.almaany.com/ar/dict/ar-en/}} to find the translation of Arabic words into English. For English-Farsi experiments, we used the Aryanpour dictionary\footnote{\url{http://www.aryanpour.com/Farsi_to_English.php}} to extract the English equivalent of Farsi words. To evaluate the performance of the classifiers we used Accuracy and F$_1$ as the previous works \cite{ji-eisenstein-2013-discriminative,agarwal2018deep,wang-etal-2016-sentence}.

For dependency parsing, UDPipe\footnote{\url{http://ufal.mff.cuni.cz/udpipe}} was used, which is a trainable pipeline for tokenization, tagging, lemmatization, and dependency parsing. We used version 2.4 of the UD pre-trained models\footnote{\url{https://lindat.mff.cuni.cz/repository/xmlui/handle/11234/1-2998}} of English, Arabic, and Farsi.

To implement the CNN network of Wang et al. \cite{wang-etal-2016-sentence}, we used the same word embedding they used. They set the size of the word vector dimension as d =300, and pre-trained the vectors with the word2vec toolkit on the English Gigaword (LDC2011T07). Hyper-parameters of the network are the same as their work.

\subsubsection{Results}
We first examine the tree kernels in the mono-lingual and then in the cross-lingual learning.

\paragraph*{Evaluation of tree-based models in mono-lingual learning}
In the first experiment, we benchmark the UD-based models on the monolingual dataset. So, we employed the original split of MSRC corpus and trained models using PT and SPT kernels. These models essentially work based on the lexico-syntactic patterns observed in training sentences. Filice et al. \cite{filice2015structural} proposed several kernels including linear, graph and SPT kernels. They showed the best accuracy is obtained using the combination of them. However, we use only tree kernels in cross-lingual experiments, to measure how much we can rely on the similarities of UD parse trees in different languages. 

As Table \ref{tb:msrc-mono} shows, tree kernels including PTK and SPTK show comparable results according to the accuracy and F$_1$ measures. This means that PT and SPT kernels, which are trained by UD parse trees, make accurate models that can be used in solving the PI task. In the next experiment, we use these models to evaluate Arabic and Farsi test data.

\begin{table}
\centering
\small
\begin{tabular}{|l|c|c|}
\hline 
Model & Accuracy & F$_1$ \\ 
\hline 
Majority Baseline & 66.5 & 79.9\\
\hline
Softmax-PTK & 76.5 & 84.1 \\ 
\hline 
Softmax-SPTK & 76.7 & 84.2 \\ 
\hline
AugDeepParaphrase \cite{agarwal2018deep} & 77.7 & 84.5 \\
\hline
Two-channel CNN \cite{wang-etal-2016-sentence} & 78.4 & 84.7 \\
\hline
$LK+GK+SM_{SPTK_{W2V}}$ \cite{filice2015structural} & 79.1 & 85.2 \\
\hline
Matrix factorization \cite{ji-eisenstein-2013-discriminative} & 80.4 & 85.9 \\
\hline 
%BiMPM \cite{Wang2017BilateralMM} (our run) & & \\
%\hline
\end{tabular} 
\caption{Results of the mono-lingual English PI using tree kernels compared with previous methods on the original split of the MSRC dataset.}
\label{tb:msrc-mono}
\end{table}

\paragraph*{Evaluation of tree-based models with UD in cross-lingual learning}
Now, we employ the parallel dataset for cross-lingual evaluation of the UD-based model trained by English data. A baseline for this task is the majority voting in that what we get if we always predict the most frequent label of the training data. A better baseline for cross-lingual PI is to use some neural models and couple them with pre-induced multilingual embeddings. So, we re-run the two-channel CNN model of Wang et al. \cite{wang-etal-2016-sentence} by our test data.

Upper bound for the cross-lingual experiment is considered the accuracy of the model when it is evaluated by the data of the same language of the training data, e.g. English. Table \ref{tab:MSRC_trainVsSemEval_LCT} shows that using PTK 61.6\% of accuracy is obtained for English test data. It is 57.7\% and 57.3\% for Arabic and Farsi, respectively; while the accuracy of the majority baseline is 50.6\%. CNN model obtained similar accuracy but much lower F$_1$ scores.

Comparing the results of Tables \ref{tb:msrc-mono} and \ref{tab:MSRC_trainVsSemEval_LCT} reveals that the accuracy of both kernels drops significantly when they are tested by our small test data. The reason is that the distribution of MSRC training data over positive and negative classes is significantly different from our test data. Specifically, 67.5\% of MSRC's training data are positive; while 50.5\% of our test data are positive.

\begin{table}
\begin{adjustbox}{width=\columnwidth}
\small
\centering
\begin{tabular}{|l|c|c|c|c|}
\hline
Model  & Training & Test  & Acc & F$_1$ \\ \hline
Baseline & English & all & 50.6 & 67.2\\ 
\hline
\multirow{3}{*}{2-channel CNN \cite{wang-etal-2016-sentence}} & \multirow{3}{*}{English} & English & 63.5 & 69.7\\
& & Arabic & 56.1  & 48.6 \\
& & Farsi & 57.3 & 58.1 \\ 
\hline
%\multirow{3}{*}{Tree-LSTM \cite{tai-etal-2015-improved}} & \multirow{3}{*}{English} & English & & \\
%& & Arabic &   &  \\
%& & Farsi &   &  \\ 
%\hline
%\multirow{3}{*}{BiPMP \cite{}} & \multirow{3}{*}{English} & English &  &  \\
%& & Arabic &  &   \\
%& & Farsi &  &  \\ 
%\hline
\multirow{4}{*}{Softmax-PTK} & \multirow{4}{*}{English}  & English & 61.6 & 69.5\\
& & Arabic & 57.7 & 68.8 \\
& & Farsi & 57.3 & 68.4 \\
& & Farsi (No UD) & 50.6 & 67.2 \\
\hline
\multirow{4}{*}{Softmax-XSPTK}& \multirow{4}{*}{English} & English & 61.0 & 69.2 \\
& & Arabic & 58.2 & 68.5 \\
& & Farsi & 58.5  & 68.9 \\ 
& & Farsi (No UD) & 50.8 & 67.2 \\
\hline
\end{tabular}
\end{adjustbox}
\caption{Results of cross-lingual PI over our split of MSRC dataset.}
\label{tab:MSRC_trainVsSemEval_LCT}
\end{table}

\paragraph*{Evaluation of tree-based models with parse formalisms rather than UD}
In this experiment, we produced dependency parse trees of Farsi data employing Hazm parser\footnote{\url{http://www.sobhe.ir/hazm/demo}} which is trained on non-UD tree-bank. Table \ref{tab:MSRC_trainVsSemEval_LCT} shows that in this case accuracy of the models significantly drops. Taking a deeper look at the tree kernels, PTK doesn't use the similarity of words and works based on exact matching of them. So, in cross-lingual experiments, it considers only the similarity of trees. In this case, accuracy on Farsi test data is 50.6\% which is the same as the majority baseline. This experiment reveals that the trees of parallel sentences that are produced by UD parsers are significantly more similar than the trees generated by other formalisms.

\subsection{Relation Extraction}

In this section, we explain the experiments of cross-lingual RE and present the results. Specifically, we compared tree-based methods including combination of tree kernels and TreeLSTM with deep methods of CNN \cite{qin2016empirical}, Bi-LSTM \cite{zhou2016attention} and RCNN \cite{lai2015recurrent}.

\subsubsection{Construction of Parallel Dataset}
SemEval 2010 released a dataset for relation extraction in task 8 \cite{hendrickx2009semeval}, which is used by many researchers. This dataset contains 8000 samples for the training and 2717 samples for the test. It was annotated with 19 types of relations: 9 semantically different relationships (with two directions) and an undirected \textit{Other} class. A brief description of these relation types is given in Table \ref{tb:rel-types}.

The SemEval-2010 dataset is in English. For cross-lingual experiments, the first 1000 samples of the test part were translated into Farsi and French. Two native Farsi and French speakers with high expertise in English were asked to translate the data.

% ----- Relation Types -------
\begin{table*}
\begin{adjustbox}{width=\textwidth,center=\textwidth}
\centering
\begin{tabu} {|l|l|}
\hline 
Relation Type & Definition \\ 
\hline 
Cause-Effect(X, Y) & \makecell[l]{X is the cause of Y, or that X causes/makes/produces/emits/... Y.} \\ 
\hline 
Instrument-Agency(X, Y) & \makecell[l]{X is the instrument (tool) of Y or, equivalently, that Y uses X.} \\ 
\hline 
Product-Producer(X, Y) & \makecell[l]{X is a product of Y, or Y produces X.} \\ 
\hline 
Content-Container(X, Y) & \makecell[l]{X is or was (usually temporarily) stored or carried inside Y.} \\ 
\hline 
Entity-Origin(X, Y) & \makecell[l]{Y is the origin of an entity X (rather than its location), and X is coming\\ or derived from that origin.} \\ 
\hline 
Entity-Destination(X, Y) & \makecell[l]{Y is the destination of X in the sense of X moving (in a physical or \\abstract sense) toward Y.}\\ 
\hline 
Component-Whole(X,Y) & \makecell[l]{X has a functional relation with Y. In other words, X has an operating or\\ usable purpose within Y.}\\ 
\hline 
Member-Collection(X, Y) & \makecell[l]{X is a member of Y.}\\ 
\hline 
Message-Topic(X, Y) & \makecell[l]{X is a communicative message containing information about Y.}\\ 
\hline 
\end{tabu} 
\end{adjustbox}
\caption{Relation types of SemEval 2010 dataset \cite{hendrickx2009semeval}.}
\label{tb:rel-types}
\end{table*}

\subsubsection{Tools and Setup}
Similar to PI's experiments, KeLP was used to implement the kernel combination. The strategy for dealing with multiple classes is ``one versus others''. For constituency parsing, Stanford CoreNLP\footnote{\url{https://stanfordnlp.github.io/CoreNLP/}} was used that contains pre-trained models for English and French within the Stanford package. For parsing Farsi data, the University of Tehran’s constituency parser\footnote{\url{http://treebank.ut.ac.ir/}} \cite{Farsiconstparser} was used. Parameter $\alpha$ of the formula \ref{eq-ck1}-\ref{eq-ck3} is 0.23 as the previous works \cite{nguyen2009convolution}. To obtain bi-lingual word embeddings, the multiCluster method of Ammar et al. \cite{ammar2016massively} was used and 512-dimensional vectors were trained for English, French, and Farsi.

\subsubsection{Result}
We first examine the tree kernels in the mono-lingual and then in the cross-lingual learning.

\paragraph*{Evaluation of tree-based models in mono-lingual learning}

There is a huge amount of works on RE, which mainly utilizes neural networks. These methods use different features including lexical, grammatical, and semantic features such as POS, WordNet, and dependency parsing.
%These methods are divided into two groups: dependency methods \cite{miwa-bansal-2016-end,cai2016bidirectional} and end-to-end methods \cite{wang-etal-2016-relation,qin2017designing}. The former uses the dependency parse tree and shortest path between two entities; while the latter consume whole sentence. 
Table~\ref{tb:pre-works} shows the state-of-the-art neural models evaluated by SemEval 2010-task 8 test set (2717 samples). The best proposed method, $CK_1$, obtained 84.0\% of F$_1$ which is comparable with the others.

%------------------------------------------------------------------------------------------------------------
\begin{table*}
\centering
\small
%\begin{adjustbox}{width=\textwidth,center=\textwidth}
\begin{tabu} spread 0,5 \textwidth{|l|l|c|}
\hline 
Classifier & Feature Set & F$_1$ \\ 
\hline 
%SVM \cite{hendrickx2009semeval} & POS, Wordnet, morphology, dependency parse, etc.  & 82.2 \\ 
%\hline 
%MV-RNN \cite{socher2012semantic} & POS, WordNet, NER  & 82.4 \\ 
%\hline 
%CNN \cite{nguyen2015relation} & - & 82.8 \\ 
%\hline 
RelEmb \cite{hashimoto-EtAl:2015:CoNLL} & embedding, dependency path, WordNet, NE  & 83.5 \\ 
\hline 
%Att-BLSTM \cite{zhou2016attention} & Position indicator & 84.0 \\ 
%\hline
$CK_1$ (Our Method) & dependency parse, constituency parse & 84.0 \\
\hline 
EAtt-BiGRU \cite{qin2017designing} & embedding, position feature & 84.7 \\ 
\hline
%CNN \cite{qin2016empirical}  & embedding, entity tag feature & 84.8 \\
%\hline
depLCNN \cite{xu-etal-2015-semantic} & dependency path, WordNet, words around nominals & 85.6 \\
\hline
BRCNN \cite{cai2016bidirectional} & dependency path & 86.3 \\
\hline
Att-Pooling-CNN \cite{wang-etal-2016-relation} & word position embeddings & 88.0 \\
\hline 
\end{tabu} 
%\end{adjustbox}
\caption{F$_1$ scores and features used by mono-lingual RE methods evaluated on the SemEval 2010-task 8 dataset.}
\label{tb:pre-works}
\end{table*}
%--------------------------------------------------------------------------------------------------
\begin{table*}
\small
\centering
\begin{tabular} {|l|c|c|c|c|c|}
\hline 
\multirow{3}{*}{\makecell[c]{Model}} & \multirow{3}{*}{\makecell[c]{Training}} & \multicolumn{4}{c|}{Test data \& size} \\ 
\cline{3-6}
& & English & English & Farsi & French \\ 
& & 2717 & 1000 & 1000 & 1000 \\
\hline 
\underline{Baselines:} & & & & & \\	
CNN \cite{qin2016empirical}  &  \multirow{3}{*}{English} & 82.8 & 82.3 & 23.8 & 46.9 \\
Att-BiLSTM \cite{zhou2016attention} & & 82.4 & 81.9 & 23.1 & 47.6 \\
RCNN \cite{lai2015recurrent} & & 82.9 & 82.3 & 28.3 & 51.1 \\
\hline
% \underline{Tree-based Models:} & & & & & \\	
Tree-LSTM & \multirow{6}{*}{English}& 79.9 & 80.0 & 52.0 & 55.6 \\

$CK_1$ & & \textbf{84.0} & 84.2 & 53.4 & 61.2 \\ 
$CK_2$ & & 79.6 & 78.8 & 65.2 & 65.2 \\ %63.6
$CK_3$ & & 83.9 & 84.0 & 59.7 & 67.5  \\ % 65.9
 
%\cline{2-13}
$CK_2$+ MWE & & 79.6 &  78.8 & \textbf{67.1} & 66.4 \\ %64.8
$CK_3$+ MWE & &  83.9 & 84.0  & 62.5 & \textbf{67.7} \\ %65.95

\hline
\end{tabular} 
\caption{F$_1$ scores of different tree-based models of RE compared to the neural models (our re-implementation) on the SemEval 2010 dataset.}
\label{tb:result}
\end{table*}

\paragraph*{Evaluation of tree-based models with UD in cross-lingual learning}

Table \ref{tb:result} shows accuracy of 84.2\% F$_1$ score for $CK_1$ when tested on the first 1000 samples of English test data. The accuracy of this model for its Farsi and French counterparts is 53.4\% and 61.2\% respectively. This kernel employs sentence context, and so it didn't show exciting results in the cross-lingual experiment; especially for Farsi data. This is because Farsi is one of the SOV languages, in contrast to English and French, which are SVO. This means verbs are usually at the end of the sentence in Farsi. When the sentence's verb is highly informative for the relation between two entities, it places outside the window surrounding two entities and so it doesn't contribute to the feature vector $V_o$.

Table \ref{tb:result} show the F$_1$ score of the models trained by $CK_2$ and $CK_3$. These kernels utilize the context words of the UD trees. Comparing three kernels, F$_1$ increased from 53.4\% to 65.2\% for Farsi, and to 67.5\% for the French test data. The best result for Farsi came from kernel $CK_2$; whereas $CK_3$ performed better with the French data. Thus, it can be concluded that the constituency-based parse trees of English and French data have more similar sub-trees than English and Farsi. The reason partially relates to the common tool for English and French; because Stanford CoreNLP has pre-trained models for both of these languages. Therefore, English and French models followed the same schema, while Farsi adopted different schema for constituency parsing.

In addition to the composite kernels, we trained a Tree-LSTM model over the UD parse trees. Tree-LSTM doesn't process the syntactic features of the input sentence, rather it takes the tokens in order of the tree's node. However, to contribute the grammatical features, for each token its word embedding was concatenated to its dependency type embedding and its POS tag embedding. The resulting network obtained 80.0\% of F$_1$ when tested by English test data. F$_1$ of this model is 52.0\% for Farsi and 55.6\% for French. 

Although the Tree-LSTM model obtained lower F$_1$ in comparison with the tree kernels, it still does better than deep baselines: we re-implemented the CNN model of Qin et al. \cite{qin2016empirical}, Att-BiLSTM of Zhou et al. \cite{zhou2016attention}, and RCNN of Lai et al. \cite{lai2015recurrent}. All networks use bilingual word embeddings in the embedding layer. As Table \ref{tb:result} shows the best F$_1$ scores were obtained by RCNN which utilizes CNN over the LSTM layer. However, the results are significantly lower than the UD-based models, specifically in Farsi. Because word order of Farsi and English sentences are very different; as Farsi is SOV and English is SVO. 

\paragraph*{Effect of Multi-Word Expressions}

\begin{figure}
\centering{
\resizebox{0.46\textwidth}{!}{
\subfloat[Farsi.\label{fig:distb-fa}]
{
\begin{tikzpicture}
\pgfplotsset{width=10cm, height = 5cm}
  \begin{axis}[ 
  ybar,
   symbolic x coords={Content-Container, Message-Topic, Cause-Effect,  Entity-Destination,  Entity-Origin,  Member-Collection, Product-Producer,   Instrument-Agency, Component-Whole},
    xtick=data,
    x tick label style={font=\normalsize,rotate=45, anchor=east},    
    nodes near coords,
    nodes near coords align={vertical},
%    ymajorgrids = true,
	ymin = 35,
	ymax = 92,
	ytick = {40, 50, 60, 70, 80}
  ]
  \addplot coordinates { (Cause-Effect, 72.5)         
  						 (Component-Whole, 52.7)
                         (Content-Container, 84.2)  
                         (Entity-Destination, 67.2)
                         (Entity-Origin,64.8) 
                         (Instrument-Agency, 60.0)
                         (Member-Collection, 62.2)
                         (Message-Topic, 76.3)
                         (Product-Producer, 61.9)
                         };
  \end{axis}
\end{tikzpicture}
}
} % end of resize box
\resizebox{0.46\textwidth}{!}{
\subfloat[French.\label{fig:distb-fr}]
{
\begin{tikzpicture}
\pgfplotsset{width=10cm, height = 5cm}
  \begin{axis}[ 
  ybar,
   symbolic x coords={Member-Collection, Cause-Effect, Message-Topic, Entity-Destination, Instrument-Agency, Product-Producer,Content-Container, Component-Whole, Entity-Origin },
    xtick=data,
    x tick label style={font=\normalsize,rotate=45, anchor=east},    
    nodes near coords,
    nodes near coords align={vertical},
%    ymajorgrids = true,
	ymin = 35,
	ymax = 92,
	ytick = {40, 50, 60, 70, 80}
  ]
  \addplot coordinates { (Cause-Effect, 75.1)         
  						 (Component-Whole, 62.7)
                         (Content-Container, 64)  
                         (Entity-Destination, 72.1)
                         (Entity-Origin, 45.1) 
                         (Instrument-Agency, 69.9)
                         (Member-Collection, 79.9)
                         (Message-Topic, 73.4)
                         (Product-Producer, 67.6)
                         };
  \end{axis}
\end{tikzpicture}
}
} % end of resize box
} % end of centring
\caption{F$_1$ scores of different classes on the Farsi and French test sets using $CK_2$ and $CK_3$ kernels respectively (the best model for each language).}
\label{fig:distb}
\end{figure}
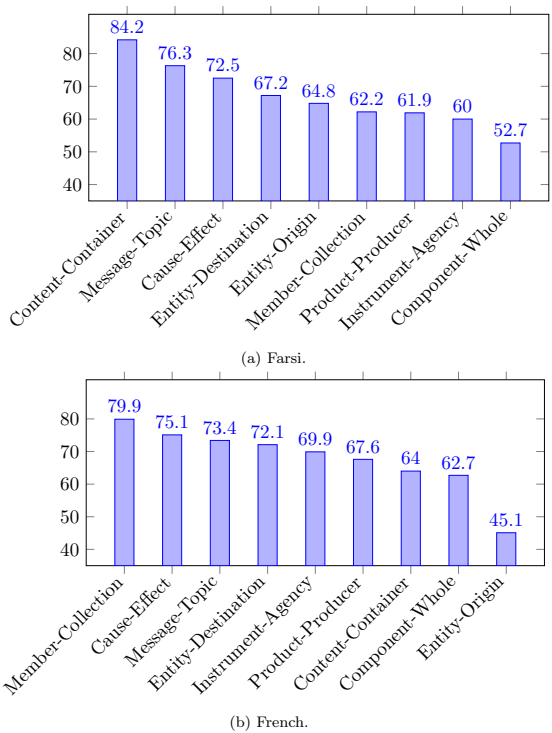

Last two rows of Table \ref{tb:result} show the F$_1$ score of the model trained on the English training data using the $CK2$ and $CK3$, in which MWEs were considered to be a single node within the dependency tree, as described at the end of Section \ref{sec:cross-re}. The accuracy of $CK_2$ mainly increased for the Farsi data, because Farsi has many multi-word expressions such as compound verbs. Farsi has only about 250 simple verbs and all the other verbs are compound \cite{samvelian2013introducing}. Considering MWE as a single node causes all the tokens which compose a verb to be treated as a single word, and so the true translation will be found when searching for that word in dictionaries. Figure \ref{fig:distb-fa} shows the F$_1$ scores of best models for different semantic classes.

\section{Discussion and Conclusion}
\label{sec:analysis}

Taking a deeper look at the proposed method, most of the mis-classifications of the cross-lingual tree models are related to the following issues:
\begin{itemize}
\item
Structural Difference: The main reason for the error of classifiers is structural differences. Although UD tries to produce as most similar trees as it can for parallel sentences, there are many language-specific dependency patterns that could not be neglected. 

\item
Lexical Gap: Words mainly convey the meaning of the sentence. A lexical gap between source and target languages usually ruins the accuracy of cross-lingual models.
\item
Confusion of different senses on a surface: Words of different languages usually have multiple senses. Confusion of different senses of words causes incorrect translation of words, because dictionaries translate word to word, but not word-sense to word-sense. On the other hand, Word Sense Disambiguation (WSD) is a difficult task and needs additional resources such as high-quality multi-lingual wordnets \cite{taghizadeh2016automatic}.
\item
Incorrect translation of prepositions: Prepositions are very informative for the RE task. Hashimoto et al. presented the five most informative unigrams and three-grams for three types of relations of the SemEval 2010-task 8 dataset \cite{hashimoto-EtAl:2015:CoNLL}, which are shown in Table \ref{tb:informative-words}. Wang et al. \cite{wang-etal-2016-relation} also presented the most representative trigrams for different relations on the same data set. Also, Lahbib et al. \cite{lahbib2013hybrid} presented the most common Arabic prepositions and showed that each one reflects some specific kinds of semantic relations. Confusion of senses for prepositions is a very common issue in word-to-word translation.
%\item
%Deletion of preposition ``of" in Farsi: In RE task, analysis showed that the preposition ``of" is important for a Component-Whole relation, but it is usually missing in Farsi. Figure \ref{fig:distb} depicts the $F1$ score of the best model for different classes of the French and Farsi test sets. As shown in Figure \ref{fig:distb-fa}, the lowest score belongs to the Component-Whole.
\item
Phrasal verbs: Phrasal verbs, which have a metaphorical meaning, often cannot be translated word for word. For example, the Farsi verb ``\FR{از دست دادن} / to give from hand'', means ``lose". When the most informative chunk of the sentence is the phrasal verb, the proposed method does not capture the true meaning. 
\end{itemize}

In general, more lexical and structural similarities between the source and target languages increase the accuracy of UD-based transfer learning. As future works, it is proposed that the UD-based approach is studied for other cross-lingual learning tasks and other languages along with different learning algorithms that are capable of dealing with parse trees.

%------------------------------------------------------------------------------------------------------------

\begin{table}
\centering
\small
\begin{tabular}{|c|c|c|}
\hline 
\makecell{Cause-Effect\\(E1, E2)} & \makecell{Content-Container\\(E1, E2)} & \makecell{Message-Topic\\(E1, E2)} \\ 
\hline 
resulted & inside & discuss \\ 
 
caused & in & explaining \\ 

generated & hidden & discussing \\ 

cause & was & relating  \\ 

causes & stored & describing  \\ 
\hline 
\end{tabular}

\begin{tabular}{|c|c|c|}
\hline 
\makecell{Cause-Effect\\(E2, E1)} & \makecell{Content-Container\\(E2, E1)} & \makecell{Message-Topic\\(E2, E1)} \\ 
\hline 
 after & full & subject \\ 
  from & included & related \\ 
 caused & contains & discussed \\ 

 triggered & contained & documented \\ 

 due & stored & received \\ 
\hline 
\end{tabular} 
\caption{Top five most informative words for some relations of the SemEval 2010-task 8 dataset \cite{hashimoto-EtAl:2015:CoNLL}.}
\label{tb:informative-words}
\end{table}

\bibliographystyle{acm}
\bibliography{SVM} 

\end{document}